# Each Prompt Matters: Scaling Reinforcement Learning Without Wasting Rollouts on Hundred-Billion-Scale MoE

Shopee LLM Team

We present CompassMax-V3-Thinking, a hundred-billion-scale MoE reasoning model trained with a new RL framework built on one principle: each prompt must matter. Scaling RL to this size exposes critical inefficiencies—zero-variance prompts that waste rollouts, unstable importance sampling over long horizons, advantage inversion from standard reward models, and systemic bottlenecks in rollout processing. To overcome these challenges, we introduce several unified innovations: (1) Multi-Stage Zero-Variance Elimination, which filters out non-informative prompts and stabilizes group-based policy optimization (e.g. GRPO) by removing wasted rollouts; (2) ESPO, an entropy-adaptive optimization method that balances token-level and sequence-level importance sampling to maintain stable learning dynamics; (3) a Router Replay strategy that aligns training-time MoE router decisions with inference-time behavior to mitigate train–infer discrepancies, coupled with a reward model adjustment to prevent advantage inversion; (4) a high-throughput RL system with FP8-precision rollouts, overlapped reward computation, and length-aware scheduling to eliminate performance bottlenecks. Together, these contributions form a cohesive pipeline that makes RL on hundred-billion-scale MoE models stable and efficient. The resulting model delivers strong performance across both internal and public evaluations.

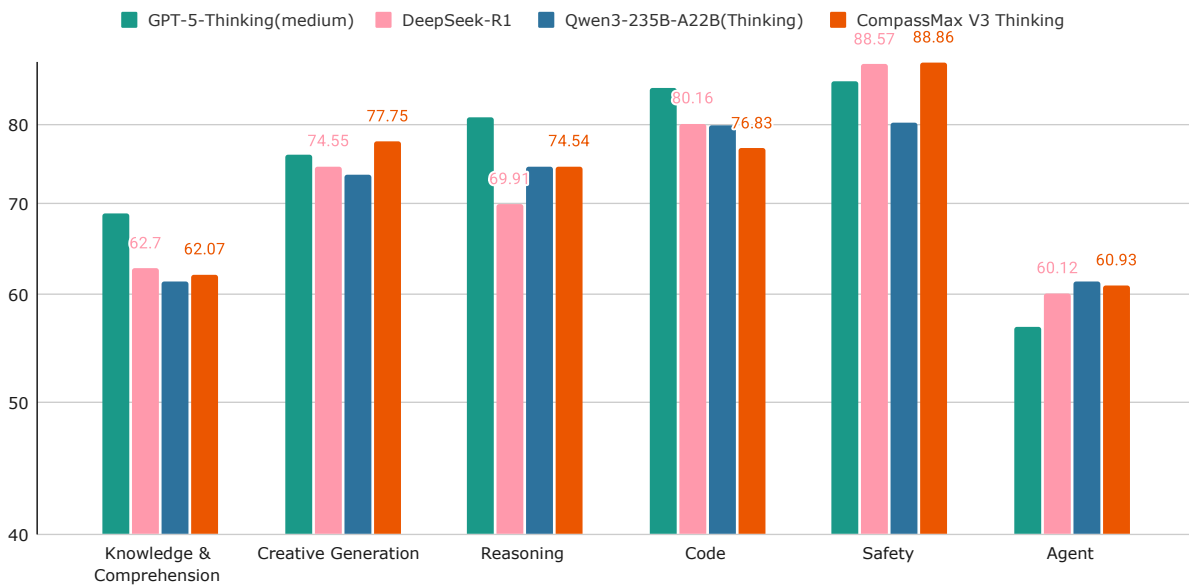

Figure 1 | CompassMax-V3-Thinking shows superior performance over competitors.

# 1. Introduction

Large-scale LongCoT models have recently demonstrated impressive progress in complex reasoning tasks (Guo et al., 2025; Team et al., 2025b), yet building a stable and efficient RL pipeline for hundred-billion-parameter Mixture-of-Experts (MoE) models remains extremely challenging. In practical deployments, several long-standing issues continue to limit performance: domain capabilities are difficult to integrate quickly, GRPO-based optimization often suffers from zero-variance prompts (Le et al., 2025a), importance sampling remains brittle at both the token and sequence levels (Zhao et al., 2025; Zheng et al., 2025), and conventional reward models can unintentionally invert advantages near the mean value of group. Furthermore, discrepancies between training rollouts and inference-time behavior frequently destabilize optimization, and multi-domain reward systems are still underexplored for long-context reasoning settings. Overall, these challenges motivate a comprehensive re-examination of how large MoE reasoning models should be optimized via RL.

A large body of prior work investigates these problems in isolation. Model merging (Ilharco et al., 2023; Yadav et al., 2023; Yu et al., 2024) and domain adaptation provide partial solutions for transferring domain knowledge. GRPO (Shao et al., 2024) and related policy optimization algorithms address reasoning capabilities but struggle when zero-variance prompts dominate the training distribution. Importance sampling corrections have been explored in sequence modeling, yet their applicability to long-horizon reasoning still has substantial room for improvement. Reward modeling has advanced rapidly, though standard BT-style (Bradley and Terry, 1952) or pointwise scoring models often exhibit non-monotonic behaviors that lead to advantage reversal. System-level optimization techniques, such as improved rollouts or more efficient reward computation, have been proposed but rarely coordinated into a unified pipeline for large-scale RL training. Yet, these solutions are often studied independently, and no prior work has unified all these aspects into a single RL framework. As a result, existing large-scale RL pipelines either suffer from one bottleneck or instability or require extensive manual tuning to keep training on track.

In this paper, we propose an integrated RL framework for large-scale MoE reasoning models that holistically addresses the above challenges through coordinated algorithm and system innovations. Our approach, used to train the CompassMax-V3-Thinking model, is guided by the principle that every prompt in training should provide a useful learning signal, thereby eliminating wasted computation. We tackle the problem on multiple fronts, with key methodological contributions summarized as follows:

- **Multi-Stage Zero-Variance Elimination:** We propose a staged mechanism to identify and suppress prompts that yield uniform (zero-variance) rewards, preventing wasted rollouts and enhancing GRPO stability. By pruning or reshaping such prompts, the policy focuses learning on diverse, informative queries—maximizing the utility of each rollout and accelerating convergence.
- **Entropy Importance Sampling Policy Optimization (ESPO):** We introduce an entropy-guided optimization method that adaptively balances token- and sequence-level importance sampling (Sheng et al., 2025). Unlike GRPO's uniform treatment, ESPO reweights updates based on entropy and reward, improving stability in long-horizon settings. By emphasizing high-entropy regions, ESPO mitigates importance sampling brittleness and implicitly prevents advantage inversion near mean-quality responses.
- **Router Replay and Advantage Stabilization:** We introduce Router Replay to align MoE expert routing between training and inference by logging and reusing gating decisions (Yao et al., 2025), reducing train–infer mismatch and improving rollout consistency. To prevent advantage inversion, we employ a Generative Reward Model (GenRM) with chain-of-thought reasoning, yielding smoother, monotonic rewards that better preserve correct advantage ordering. Together, these methods enhance RL stability for large-scale MoE training.

- **High-Throughput Rollout System:** We design a high-efficiency RL training system to eliminate system bottlenecks in scaling long-context MoE models. Key components include: (a) FP8-precision rollout for faster inference; (b) length-aware scheduling to reduce stragglers and improve GPU utilization; and (c) overlapped reward computation with token generation to minimize idle time. These optimizations jointly enable practical, scalable RL training with high throughput and minimal performance trade-offs.

Together, these contributions form a cohesive RL pipeline that enables hundred-billion-parameter MoE models to learn complex reasoning tasks stably and efficiently at scale. Unlike prior approaches that addressed each pain point in isolation, our unified framework tackles algorithmic stability and system performance in concert. This end-to-end design provides both a practical recipe and a conceptual blueprint for training high-performance LongCoT RL systems. We demonstrate that with careful elimination of wasted prompts, adaptive optimization techniques, alignment of training with inference conditions, and system-level optimizations, reinforcement learning can be successfully scaled to previously infeasible model sizes. In contrast to fragmented solutions in the literature, CompassMax-V3-Thinking exemplifies a methodological advance where all pieces of the RL puzzle—reward modeling, policy optimization, and engineering—are jointly optimized. This integrated approach makes it possible to train massive MoE models without RL collapse or inefficiency, marking a significant step toward truly scalable long-context reasoning.

## 2. Methods

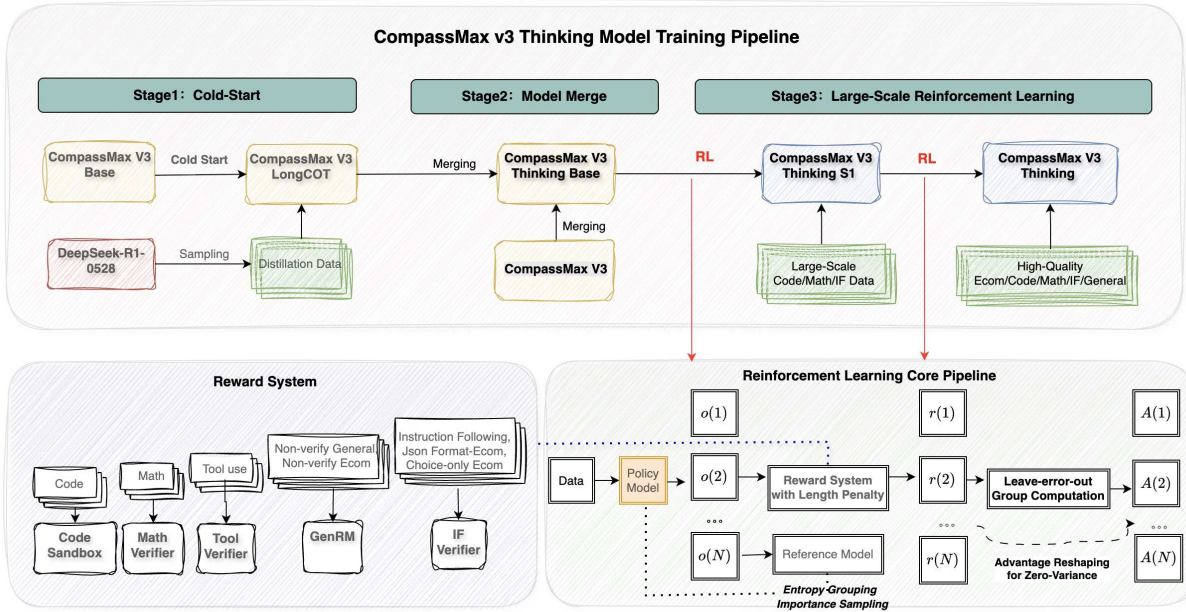

Figure 2 | Overview of **CompassMax-V3-Thinking Model Training Pipeline.** The pipeline consists of three stages: (1) Cold-Start SFT using Long-CoT data distilled from DeepSeek-R1-0528; (2) Model Merging with CompassMax-V3 SFT checkpoints to produce the Thinking Base; (3) Two-phase Large-Scale RL to refine structured reasoning (code, math, IF) and expand domain skills (e-commerce, tool-use, QA, general reasoning).

## 2.1. Training Pipeline of CompassMax-V3-Thinking

The pipeline proceeds in three major stages: (1) **Cold-Start** via Long-CoT supervised fine-tuning, (2) **Model Merge** to integrate complementary SFT capabilities, and (3) **Large-Scale Reinforcement Learning (RL)** to refine reasoning correctness and domain generalization. During the reinforcement learning (RL) phase, the training process is divided into two sequential phrases, each designed to progressively expand the model's reasoning scope and domain adaptability while maintaining learning stability. In the first stage, we perform large-scale RL training primarily on datasets covering **code generation**, **mathematical reasoning**, and **instruction following**. These domains provide a rich foundation for structured, step-by-step reasoning and logical consistency, which are essential for developing a strong "thinking" capability. The objective at this stage is to optimize the model's ability to decompose complex tasks, follow instructions accurately, and produce verifiable outputs across formal reasoning domains. We refer to this phase as the *reasoning RL stage*. After the reasoning RL phase stabilizes, we extend the training domains to include **e-commerce**, **tool-use**, and **general question answering (QA)**. These additional datasets enable the model to generalize beyond structured reasoning tasks into more dynamic, multi-context environments that require contextual understanding and decision grounding. To maintain training efficiency and prevent catastrophic forgetting, we retain only samples from the previous stage that fall within the *medium-to-high difficulty range*. This selective sampling ensures that the model continues to receive sufficiently challenging reasoning stimuli without being dominated by trivial examples, thereby reinforcing both adaptability and robustness.

## 2.2. Long-CoT Supervised Fine-Tuning

The cold-start phase plays a crucial role in determining both the convergence speed of reinforcement learning (RL) and the final generalization ability of the model. An improper initialization or biased data distribution can lead to overfitting within specific domains and hinder the learning of broad reasoning patterns. To mitigate these issues, we design a carefully controlled cold-start data pipeline.

We first collect a diverse set of **multi-domain prompt data** from external sources, covering a wide range of reasoning categories, including mathematics, code, e-commerce, dialogue, and multilingual understanding. This heterogeneous dataset provides a broad contextual foundation for subsequent policy optimization.

Next, we perform **knowledge distillation** from high-capacity large language models (LLMs) to transfer advanced reasoning and problem-solving behaviors into the base model. To ensure both safety and data quality, a series of filtering procedures are applied:

- **Deduplication:** Removing duplicated or near-duplicated samples to prevent distributional bias and overfitting.
- **Test-set filtering:** Excluding any data overlapping with evaluation benchmarks to avoid data leakage and inflated performance.
- **Harmful content filtering:** Detecting and discarding prompts or responses containing toxic, biased, or unsafe content.

This multi-stage process ensures that the RL training starts from a robust and diverse initial distribution, improving both convergence stability and generalization across unseen domains. By combining multi-domain coverage with distilled reasoning priors, the model benefits from a strong cold-start foundation, leading to faster adaptation and more consistent downstream performance.

The base model is fine-tuned on Long-CoT SFT data. This produces a checkpoint, *CompassMax-V3-Thinking LongCoT*, capable of long reasoning in multiple domains.

The Cold-Start stage bootstraps explicit reasoning traces, enabling later verifier-based RL rewards and improving interpretability.

## 2.3. Model Merging for Rapid Capability Acquisition

In our e-commerce domain, training a domain-specific model with extensive long Chain-of-Thought (long-CoT) reasoning would require substantial computational resources, which are far beyond our current capacity. Moreover, there are no existing long-CoT models tailored for the e-commerce domain that could be directly used for distillation. To overcome this limitation, we adopt a model merging approach as an efficient alternative to full-scale long-CoT training.

In industry practice, model merging is often applied between different training stages of the same model to mitigate overfitting and improve generalization. In contrast, our approach merges a general multi-domain long-CoT model with 3 model, which is highly optimized for e-commerce tasks, enabling us to transfer reasoning capabilities without incurring large-scale training costs.

We systematically evaluated several model merging techniques, including TA (Ilharco et al., 2023), DARE (Yu et al., 2024) and TIES (Yadav et al., 2023), and found that the **TIES** strategy delivered the most consistent performance. We attribute this to its mechanism of eliminating noisy weight directions during optimization, resulting in cleaner and more stable merged representations.

Furthermore, our experiments revealed that, during merging, the weight of the long-CoT model affect the merging result significantly. This work represents the **first systematic exploration** of model merging between distinct reasoning modes in large-scale **MoE architectures**, providing both empirical insights and practical parameter settings for future applications in this domain.

In our model merging strategy, we used a cost-efficient method to directly integrate the e-commerce knowledge from CompassMax-V3 into the CompassMax-V3-Thinking model after its cold-start phase.

## 2.4. Large-Scale Reinforcement Learning

For the efficiency of convergence in reinforcement learning, we introduce a set of optimizations that addressed several core issues, including two parts. **Speeding of Convergence**: the *zero-variance reward problem* within GRPO-based RL, the *uniform importance sampling ratio* issue. **Stability of Convergence**: The *mismatch between training and inference*. We also mitigated the *advantage flipping* when reward near the mean value, enabling the model to train on larger datasets and maintain longer training steps with stable performance.

### 2.4.1. Multi-Stage Zero-Variance Elimination

In reinforcement learning (RL) for large language models, a common issue we observe is the **zero-variance prompt (ZVP)** problem. It occurs when, within a GRPO group, all rollouts receive identical rewards—typically in the early stage when the policy is too weak and all responses are poor, or in the late stage when the policy is too strong and all responses are good. In such cases, the within-group reward variance becomes zero, leading to zero advantage estimates and thus no effective gradient signal. This results in wasted computation and hampers convergence.

Existing industrial approaches, such as *dynamic sampling*, attempt to alleviate this issue by discarding zero-variance groups. However, this strategy further exacerbates computational inefficiency, as large portions of rollouts are ignored.

To address this, we propose a unified strategy to reduce the impact of ZVP across all RL pipelines by (1) **expanding the exploration space**, (2) **reshaping rewards**, and (3) **enhancing advan-** **tage estimation**. These complementary methods jointly maintain sufficient variance during policy optimization, thereby accelerating convergence and improving training stability.

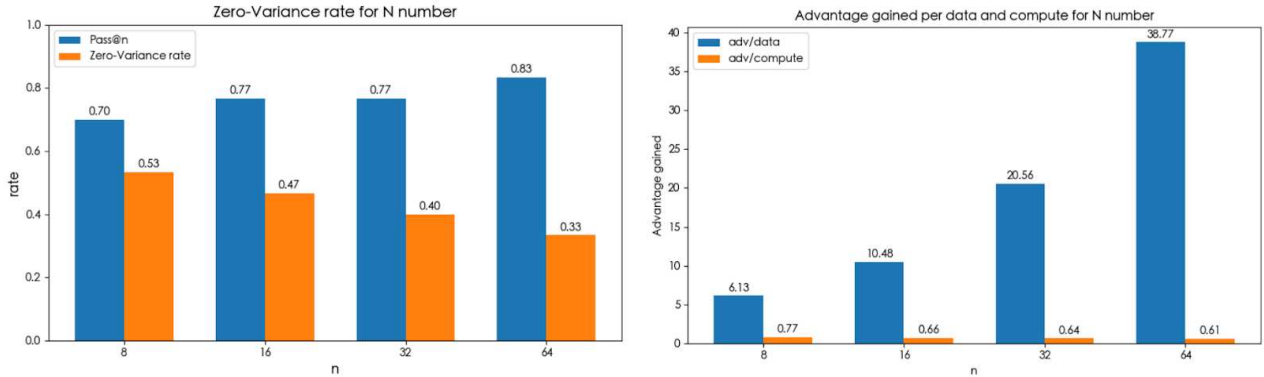

Figure 3 | Left: Zero-variance Rate decrease with larger *N* while pass@n grows; Right: The advantage absolute value of each unit of data and computing resources.

- **Rollout Stage: Expanding the exploration space.** While many existing works rely on dynamic filtering (Yu et al., 2025) to improve rollout quality, we regard rollout computation as highly valuable—especially in large-scale MoE models where discarding rollouts is costly. Instead of filtering out large portions of data, we explored methods to directly reduce the zero-variance rate. From a data perspective, we first filtered out samples that consistently received high validation pass rates from smaller external models, as these tend to produce uniformly perfect rewards. To promote greater reward diversity, we increased the model's exploration space by tuning the sampling size *N* for the Max model and continuously monitored the zero-variance ratio. Experimental results show that expanding the exploration space reduced the overall zero-variance rate by **17%**. Based on the trade-off between computational cost and marginal gain, we selected an optimal *N* size for continued training.
- **Reward Stage: Reward reshaping.** We redefined reward computation to rely primarily on *pass rates*—for example, those from code sandbox or instruction-following verifier. Additional penalty terms, including *length penalties* and *repetition penalties*, were introduced to reshape the reward landscape. These auxiliary penalties effectively leverage negative samples, especially in cases where the model engages in overly long reasoning sequences that exceed its capability, thereby enhancing negative advantage learning.
- **Actor Update Stage: Advantage reshaping.** Even after the above optimizations, approximately 5% of sample groups still exhibited zero-variance rewards. To address this, we integrate RL-ZVP (Le et al., 2025b) to reshape advantage, which injects controlled stochasticity into the advantage estimation process. This further stabilizes learning dynamics and prevents collapse in cases of extremely uniform rewards.

### 2.4.2. ESPO: Entropy Importance Sampling Policy Optimization

As discussed in the *GSPO* paper, the inconsistency between training and inference in MoE models under the GRPO framework often causes dramatic fluctuations in importance sampling ratios, leading to unstable learning dynamics. **GSPO tackles this by applying a *sequence-level* importance sampling rate, where all tokens within a sequence are treated equally.** However, this contradicts findings from 80/20 rule-based experiments (Wang et al., 2025) on dense models, where incorporating approximately 20% of high-entropy tokens yields better exploration and stability. We reproduced GSPO's results and confirmed that a uniform treatment of tokens limits optimization efficiency.

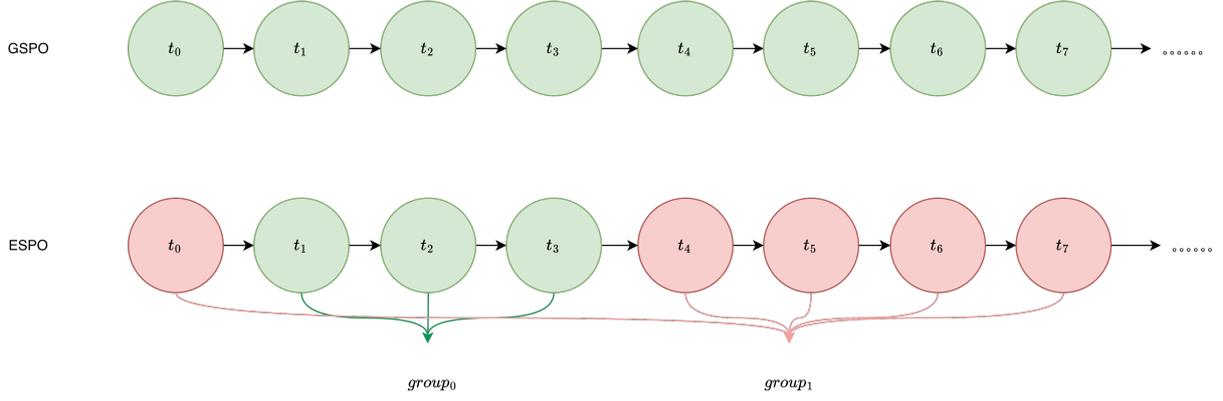

Figure 4 | The importance sampling ratio of ESPO compare to GSPO. GSPO using the same importance sampling ratio for every token, while ESPO make it different within each token groups.

Based on these insights, we propose the **ESPO** which achieves a more balanced trade-off between token-level and sequence-level importance sampling.

ESPO decomposes each sequence into entropy-coherent token groups and performs policy optimization at the *token-group level*. This design allows adaptive credit assignment across token groups, improving gradient efficiency without sacrificing stability. ESPO defines the following objective:

$$\mathcal{J}_{\text{ESPO}}(\theta) = \mathbb{E}_{x \sim \mathcal{D}, \{y_i\}_{i=1}^G \sim \pi_{\theta_{\text{old}}}(\cdot|x)} \left[ \frac{1}{G} \sum_{i=1}^G \frac{1}{|\tau|} \sum_{\tau} \frac{1}{|y_\tau|} \sum_{t=1}^{|y_\tau|} \min\left( s_\tau^{i,t}(\theta) \hat{A}_\tau^{i,t}, \text{clip}(s_\tau^{i,t}(\theta), 1 - \epsilon_\tau, 1 + \epsilon_\tau) \hat{A}_\tau^{i,t} \right) \right],$$

where $s_\tau^{i,t}(\theta)$ denotes the token-level importance ratio within group $\tau$, $\epsilon_\tau$ is the entropy-adaptive clipping threshold, and $\hat{A}_{i,t}$ is the normalized token-level advantage.

- **Entropy Grouped Importance Sampling:** Within each sequence, tokens are grouped by their entropy values. High-entropy tokens—which represent areas of greater uncertainty or exploratory potential—are assigned independent importance sampling rates, rather than being averaged together with low-entropy tokens. This prevents under-updating of exploratory regions and promotes more efficient policy learning. we define the group-wise importance ratio as follows:

$$s_\tau^{i,t}(\theta) := sg\left[ s_\tau^i(\theta) \right] \cdot \frac{\pi_\theta(y_{i,t} | x, y_{i,<t})}{sg\left[ \pi_{\theta_{\text{old}}}(y_{i,t} | x, y_{i,<t}) \right]} \quad (1)$$

where $sg[\cdot]$ denotes the stop-gradient operator that blocks backpropagation through the detached component. The group-level scaling term $s_\tau^i(\theta)$ is defined as the length-normalized sequence ratio:

$$s_\tau^i(\theta) = \left( \frac{\pi_\theta(y_\tau | x)}{\pi_{\theta_{\text{old}}}(y_\tau | x)} \right)^{\frac{1}{|y_\tau|}} \quad (2)$$

- **Entropy-adaptive Clipping:** ESPO uses entropy statistics to dynamically compute each group's clipping bound:

$$\epsilon_\tau = \frac{\alpha}{|\tau|} \sum_{t \in \tau} \frac{e_t^i}{\log |\mathcal{V}|} \quad (3)$$

where $\alpha$ is a global scaling factor and the denominator $\log |\mathcal{V}|$ serves as the theoretical upper bound of token entropy, normalizing the clipping coefficient across different vocabulary scales.

### 2.4.3. Preventing advantage flipping

In group-based advantage computation for RL, reward models trained with the *Bradley–Terry (BT) loss* often exhibit nonlinear reward curves, making group samples near the mean reward value incomparable. This can lead to *advantage flipping*, where the advantage direction is incorrectly computed. To address this, we trained a **GenRM (Generative Reward Model)** incorporating *Chain-of-Thought (CoT)* reasoning, which also demonstrated higher accuracy than ORM baselines. Furthermore, we introduced a **three-class comparison** scheme—categorizing outputs as "better," "similar," or "worse" relative to reference answers—so that the GenRM can correctly assign reward scores below or equal to the reference level. This design fundamentally prevents advantage flipping by ensuring consistent gradient direction.

### 2.4.4. Router Replay for Training-Inference Mismatch

Modern RL frameworks often adopt separate architectures for the inference and training stages. For instance, vLLM is typically used for rollout generation, whereas Megatron handles the training process. This decoupled design, however, can introduce inconsistencies in token probability distributions between the two phases due to underlying implementation differences and floating-point numerical errors, occasionally resulting in catastrophic RL collapse. Several prior works have addressed this issue from different perspectives. (Liu et al., 2025; Team et al., 2025a; Yao et al., 2025) proposed corrections or adjustments to importance sampling or policy gradients, which can temporarily delay training collapse but do not fully resolve the underlying instability. (He and Lab, 2025) attributed the instability to nondeterministic behavior in inference engine kernels and introduced batch-invariant kernels; however, this approach incurs substantial performance overhead. Another line of work (Qi et al., 2025) identified BF16 numerical precision as a primary source of training–inference mismatch and suggested replacing BF16 with FP16. In our experiments, we observed that for extremely large models, FP16 computations suffer from severe overflow, limiting the applicability of previous precision-based solutions.

To address this issue, we analyzed discrepancies between training and inference log-probabilities and found that these discrepancies become particularly pronounced after the MoE router. By aligning router selections between training and inference, the differences were significantly reduced. Based on this observation, **we developed a Router Replay mechanism for RL training**. Specifically, router decisions for all tokens are recorded during the vLLM rollout phase and directly reused when Megatron recomputes log-probabilities. This approach corrects the mismatch between training and inference log-probability distributions, reducing the discrepancy from the order of $10^{-3}$ to $10^{-4}$ and effectively stabilizing RL training. Notably, **our findings are consistent with concurrent work by Xiaomi** Ma et al. (2025).

### 2.5. Compass-Gym: A Multi-domain Reward System

We built a multi-domain **reward system** for our reinforcement learning framework. While VeRL provides a basic **math verifier** and **code sandbox** component, we developed *early stop*, *failure retry mechanisms* and *concurrency optimization* to improve robustness and throughput of code sandbox, **raising QPS to 3.5x faster**. The reward of code is computed as the execution success rate, which mitigates sparse reward issues by providing dense feedback from internal test results. This design reduces variance within code-based reasoning trajectories and stabilizes optimization. Our system also incorporates multiple domain-specific enhancements to support other RL training scenarios.

We firstly develop an **instruction following verifier** in Compass-Gym, which supports over tens of instruction constrains including keyword detection, word or symbol counting, response formating, language detection and so on. Tool-use, GenRM are also integrate to our Compass-Gym making complete reward signals for RL. Finally, the first reward system for e-commerce is design to enhance the e-commerce capabilities of CompassMax-V3-Thinking.

### 2.5.1. Tool-Use Reward

We train the model with a structured reward for tool-integrated reasoning. Following prior work on ToolRL Qian et al. (2025), we decompose the per-turn reward into a *format* term that enforces a strict output schema and a *correctness* term that assigns dense, partial credit for the content of tool calls:

$$R_{\text{final}} = R_{\text{format}} + R_{\text{correct}}. \tag{4}$$

**Format reward.** We use a minimal, order-aware binary check:

$$R_{\text{format}} \in \{0,1\} = \begin{cases} 1, & \text{if the trace is well-formed and fields appear in the required order;} \\ 0, & \text{otherwise.} \end{cases} \tag{5}$$

Concretely, we require `<think>` to appear, and then either a (possibly empty) sequence of `<tool_call>` JSON objects or a text response.

**Correctness reward.** Let $G = \{G_1, \ldots, G_n\}$ be the set of ground-truth tool calls for the current turn and $P = \{P_1, \ldots, P_m\}$ the model's predicted calls. We score along three axes and then compute an optimal assignment between predicted and gold calls to maximize the total score.

**(a) Tool-name overlap.** Let $N_G$ and $N_P$ denote the sets of tool names in $G$ and $P$, respectively. We define a set-overlap term

$$r_{\text{name}} = \frac{|N_G \cap N_P|}{|N_G \cup N_P|} \in [0,1]. \tag{6}$$

**(b) Parameter-key overlap.** For a predicted–gold matching, let $\text{keys}(\cdot)$ extract parameter *names*, let $\text{keys}(P_G)$ and $\text{keys}(P_P)$ denote the parameter name of the ground truth function call and predicted function call. We sum Jaccard overlaps of parameter keys for each matched pair:

$$r_{\text{param}} = \sum_{G_j \in G} \frac{|\text{keys}(P_G) \cap \text{keys}(P_P)|}{|\text{keys}(P_G) \cup \text{keys}(P_P)|}, \quad r_{\text{param}} \in [0, |G|]. \tag{7}$$

**(c) Parameter-value matches.** For each matched pair and key $k$ in the gold call we add one point for an exact value match:

$$r_{\text{value}} = \sum_{G_j \in G} \sum_{k \in \text{keys}(G_j)} \mathbf{1}[P_G[k] = P_P[k]], \quad r_{\text{value}} \in \left[0, \sum_{G_j \in G} |\text{keys}(G_j)|\right], \tag{8}$$

where $P_G[k]$ and $P_P[k]$ denotes the values of the ground truth and predicted parameter $k$.

**Optimal assignment and normalization.** Let $r_{\text{match}} = r_{\text{name}} + r_{\text{param}} + r_{\text{value}}$ and let $R_{\text{max}}$ be the maximum attainable total by an optimal bipartite matching between $P$ and $G$. We normalize with

$$S_{\text{max}} = 1 + |G| + \sum_{G_j \in G} |\text{keys}(G_j)|, \tag{9}$$

and define

$$R_{\text{correct}} = 6 \cdot \frac{R_{\text{max}}}{S_{\text{max}}} - 3 \in [-3, 3]. \tag{10}$$

This yields dense feedback for (i) choosing the right tools, (ii) populating the correct parameter *schema*, and (iii) filling *values* correctly, while penalizing missing or spurious calls via the matching.

### 2.5.2. Generative Reward Model (GenRM)

The goal of this part is to build a Generative Reward Model (GRM) with both reasoning and evaluative capabilities. Recent studies have shown that effective reward models should not merely decide which answer is better, but also understand why —that is, they must possess a certain degree of reasoning and self-consistency. Based on this insight, we choose DeepSeek-Distilled-Qwen3-8B-0528 as our base model, as it demonstrates strong chain-of-thought reasoning and robust decision-making abilities, providing a solid foundation for interpretable and generalizable reward modeling.

Our training approach follows the principles of RM-R1 (Chen et al., 2025), but introduces a key structural extension to address its inherent limitation. The original RM-R1 formulation performs binary preference classification (*A better than B*), which cannot express ties between responses of similar quality. To overcome this, we expand the decision space such that the GRM produces a ternary label

$$\hat{l} \in \{A, B, TIE\},$$

and outputs it at the end of a structured reasoning chain. Specifically, the model first generates intermediate reasoning fields such as <type>, <rubric>, and <eval>, representing the judgment type, evaluation rationale, and fine-grained assessment, respectively, and finally outputs the conclusive label within the <answer> field. The learning objective is correspondingly extended from binary correctness to ternary correctness:

$$\max_{r_\theta} \mathbb{E}_{(x, y_a, y_b, l) \sim D, \hat{l} \sim r_\theta} [\mathbf{1}[\hat{l} = l]],$$

which encourages the model not only to select the correct label but also to learn how to justify its choice through structured reasoning, achieving a closed-loop of generation, evaluation, and decision.

To preserve general capability, we retain the general preference data used in RM-R1 while adding 19K e-commerce domain samples to enhance performance on product title rewriting, brand recognition, and related tasks. In addition, we introduce 6K tie-labeled samples to strengthen the model's ability to recognize TIE cases across multilingual and domain-diverse scenarios, ensuring stable behavior when the quality difference between responses is subtle.

We first train the GenRM with GRPO on a fixed preference dataset and freeze it before starting CompassMax-V3-Thinking RL. In the Compass Thinker reinforcement learning stage, when a reference answer is available, the GRM serves as the reward signal generator: if the model output is judged as better than or equal to the reference, a reward of 1 is given; if it is worse, the reward is 0. This generative, self-judging reinforcement mechanism enables a reasoning–generation–evaluation loop that consistently refines preference alignment and interpretability.

On Shopee's in-house evaluation set, the **GRM** significantly outperforms the baseline model. Its agreement rate with GPT-4 judgments reaches 84.3%, exceeding the baseline ORM's 74.1% by 9.4 percentage points. Detailed results show that GRM maintains high accuracy across all response-quality categories (*much better*, *better*, *about equal*, *worse*, *much worse*), achieving an overall accuracy of 83.8% on this benchmark—substantially higher than the 64.3% accuracy of the Skywork-v2-based 8B ORM used in CompassMax-V3 (Maria, 2025). Its performance on tie recognition is particularly strong (98.8%), far surpassing the ORM's 51.2%. These results demonstrate that introducing structured reasoning and ternary decision modeling yields more stable and fine-grained preference assessments.

*2.5.3. E-commerce Reward System*

To enhance the e-commerce capabilities of CompassMax-V3-Thinking, we design the first reward system tailored to Shopee's e-commerce domain. This system extends the generic Code/Math/Tool/Instruction rewards with domain-specific verifiers and judges, and is used on both public e-commerce datasets and Shopee in-house tasks across multiple languages (English, Chinese, Thai, and other Southeast Asian languages).

We implement three complementary reward components:

1. **Keyword-based verifier for multiple-choice questions.** Many e-commerce tasks can be cast as discrete selection: the model chooses one option (e.g., an item, category, label, or brand) from a finite candidate set. We convert such tasks into multiple-choice questions and design a multilingual, keyword-based verifier that maps the model's answer back to a canonical option string. The reward is 1 if the predicted option matches the ground-truth label and 0 otherwise.
2. **Instruction-following verifier with JSON schema checks.** For extraction-style tasks, we require the model to output a structured JSON object following a strict schema. The instruction following verifier deterministically checks (i) JSON validity, (ii) presence of required keys, and (iii) lexical or ID-based correctness for each field. Field-level pass/fail signals are aggregated into a dense score in $[0, 1]$, which we rescale to the target reward range (e.g., $\{0, 1\}$ or $[-1, 1]$).
3. **Generative Reward Model (GenRM) for unstructured text answers.** For open-ended e-commerce tasks where labels are not uniquely defined, we rely on the Generative Reward Model (Sec. 2.5.2) as a domain-aware judge. GenRM compares the policy output against a reference answer and produces a ternary verdict (worse/tie/better), which is converted into a scalar reward (e.g., reward 1 if better or equal to the reference, and 0 otherwise) for GRPO.

These three components provide flexible and robust reward signals for a wide spectrum of e-commerce tasks. On open-source datasets, we apply them to tasks such as: *next purchase item prediction, similar brand retrieval, similar product retrieval, sentiment analysis of user reviews, product attribute extraction, package weight estimation, and brand extraction*. For Shopee in-house scenarios, we extend the same mechanisms to tasks including: *brand and sub-brand extraction, address parsing (floor/room/street/subdistrict/district), selling point extraction and migration, item category classification, query–item relevance judgment, fine-grained product attribute extraction, and similar brand/product retrieval*.

Overall, the e-commerce reward system bridges programmatically verifiable signals (keyword and JSON verifiers) with GenRM-based judgments for open-ended outputs, providing dense and domain-aligned rewards for real-world e-commerce workflows.

## 3. System-Level Training Efficiency Optimization

While algorithmic improvements enhance the effectiveness of policy learning, the overall throughput of reinforcement learning (RL) training is often constrained by system-level bottlenecks. With increasing sequence lengths in reinforcement learning (RL), the rollout phase has begun to dominate overall training time — a fact corroborated by the profiling results shown in Figure 5. Consequently, optimising the rollout stage has become a key lever for improving end-to-end training efficiency. To this end, we propose four complementary optimisation directions: **In the rollout stage**, we introduce two key optimisations: a **length-based load balancing** strategy at the scheduling layer to mitigate the long-tail effect caused by variable generation lengths across distributed workers, and an **FP8-quantised rollout** method to accelerate inference and shorten rollout latency. **In the reward computation stage**, we apply two complementary techniques: **multi-detokenisation parallelism**, which alleviates tokeniser-induced performance bottlenecks, and **reward overlap**, which overlaps reward computation with rollout generation — computing rewards immediately once each rollout sample is completed — to further improve pipeline efficiency and reduce idle GPU time.

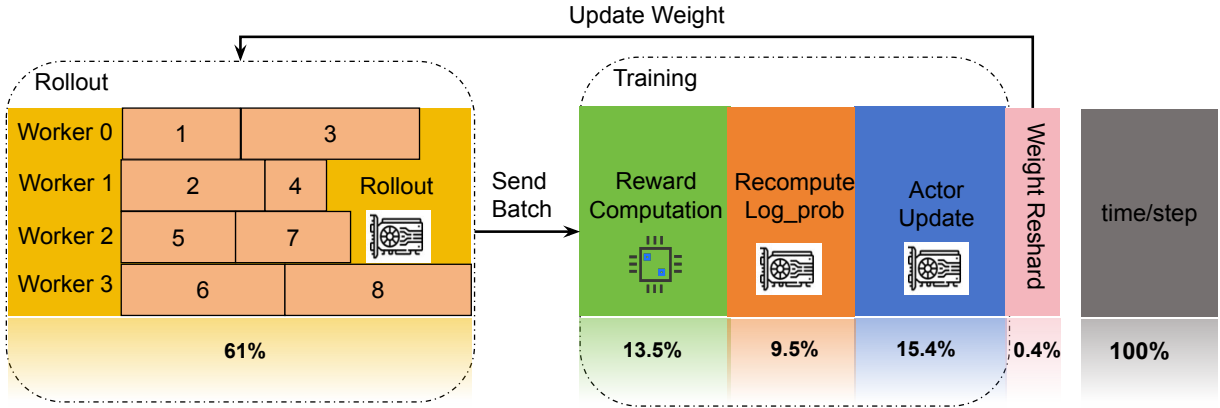

Figure 5 | Overview of **Stage-wise Time Consumption in the profiling.** The figure presents profiling results for the RL training of the CompassMax-V3-Thinking, using 256 × H100s, with rollout input length of 1k tokens, output length of 32k tokens, batch size = 512 and n = 8. It is evident that the rollout stage constitutes the majority of the runtime, followed by actor update and reward computation.

Table 1 | The experimental results of efficiency optimization methods.

| Method | Overall Speedup ↑ |
| --- | --- |
| Base line | 1.00× |
| + Multi-detokenisation parallelism | 1.16× |
| + Reward computation overlap | 1.17× |
| + FP8-quantised rollout | 1.52× |
| + Length-Based load balancing | 1.66× |

### 3.1. Optimization for Rollout

#### 3.1.1. Length-Based Load Balancing for Rollout Synchronization

In a synchronous rollout setup based on *Ray* or similar actor frameworks, each worker group must complete its assigned trajectories before the system can proceed to the next iteration. However, due to the presence of long-tail samples—where a few extremely long sequences dominate computation

time—the synchronization barrier often forces faster workers to idle while waiting for slower ones to return results through `ray.get()`. This leads to significant under-utilization of compute resources and prolonged rollout latency.

Industrial solutions typically rely on complex asynchronous training architectures or adaptive job schedulers to mitigate this problem, but such systems incur nontrivial engineering cost and consistency challenges. Instead, we adopt a lightweight strategy that leverages length prediction for **pre-rollout load balancing**.

Before rollout begins, we use the base model to generate a short *draft response* for each input, allowing us to estimate its expected generation length. Based on these predicted lengths, we reorder and distribute samples such that each worker group receives a batch with approximately equal total expected decoding length. This ensures that workers finish their rollouts within similar time windows, substantially reducing the variance in completion times.

This method preserves the synchronous training paradigm and requires only a few lines of preprocessing code. Despite its simplicity, it leads to a measurable system throughput improvement: in internal benchmarks, overall rollout speed increased by approximately **8%**, and the GPU idle ratio dropped by over 12%. The technique effectively achieves the benefits of asynchronous scheduling while maintaining the simplicity and determinism of synchronous rollout execution.

### 3.1.2. FP8 Quantisation for Rollout

The aforementioned length-based load balancing optimisation can thus be viewed as a scheduling-level strategy to streamline the rollout process, improving GPU utilisation and reducing overall time. To further optimise the rollout itself, the rollout engine (e.g., vLLM) supports an efficient FP8 quantisation scheme for weights and activations on modern hardware such as the NVIDIA H100. By quantising weights to FP8 representation, throughput is significantly increased—up to $1.6\times$ according to the vLLM documentation—without additional computational resources, thereby reducing generation token-processing (TP) time and accelerating the rollout stage.

To implement the FP8-based rollout, we first enabled the FP8 quantisation parameters of the vLLM engine. Secondly, we extended the weight update function for rollout engine in our training engine, enabling block-wise quantisation of weights. This ensured that the rollout engine could correctly load the quantised weights. In parallel, we modified vLLM weights process methods for Linear and MoE layers by retaining the weight loader of FP8 quantized weights, thereby satisfying the requirement for repeated weight updates over multiple RL iterations.

Through the this optimisations, the FP8-quantised rollout achieved a 30% reduction in rollout time under a generation length of 32k tokens, and the end-to-end overall training time was reduced by nearly 20%.

## 3.2. Optimization for Reward Computation

### 3.2.1. Multi-Detokenization for Reward Processing Acceleration

Another major bottleneck in the reward annotation stage arises from the detokenization of large volumes of generated text before scoring by the reward model. In reasoning-oriented rollouts with large $N$, detokenization can account for up to 20–30% of total pipeline latency. To mitigate this, we implement a **multi-detokenize** mechanism that parallelizes detokenization across multiple CPU workers, aggregating results asynchronously while preserving interface compatibility. This multi-process concurrency fully utilizes host resources, improving reward labeling throughput by an average of 14%, with greater gains observed in long-form reasoning rollouts, without affecting rollout semantics or policy reproducibility.

### 3.2.2. Reward Computation Overlap

Due to the presence of a long-tail effect in the rollout stage, waiting until the completion of all rollouts before beginning reward computation clearly slows down end-to-end training efficiency. To address this issue, we shift the reward calculation to begin immediately upon completion of each individual rollout response. This optimisation – overlapping reward computation with the rollout phase – reduces the time spent in the reward stage by approximately 85%.

## 4. Evaluation

We evaluate CompassMax-V3-Thinking across four axes: (1) an in-house e-commerce benchmark covering realistic SEA business scenarios, (2) in-house multilingual datasets spanning seven SEA languages, (3) a general-ability battery aggregated from internal tasks, and (4) the ARC(Agent, Reasoning, Code) ability evaluation with open-source benchmarks. Our in-house evaluation framework builds on the `llm-evaluation-harness` (Gao et al., 2024) for standard capabilities and uses GPT-4.1 as an adjudicator for open-ended, domain-specific prompts to ensure fairness and consistency.

### 4.1. In-House Dataset Construction

Shopee operates across more than ten markets in Southeast Asia and beyond (e.g., Singapore, Malaysia, the Philippines, Thailand, Vietnam, and Brazil). Because our products and users are rooted in e-commerce, *domain-specific* capabilities are our primary focus. Yet most public LLM evaluations emphasize general abilities, which do not faithfully capture performance in e-commerce workflows. However, benchmarks derived from *real* e-commerce business scenarios (e.g., product guidance, after-sales service, and product understanding) remain scarce. In addition, existing public datasets are largely limited to English and Chinese, with little to no coverage of Southeast Asian languages. To close these gaps, we build SEA e-commerce evaluation sets directly from realistic business scenarios, covering seven languages: English(en), Traditional Chinese(tw), Indonesian(id), Vietnamese(vi), Thai(th), Portuguese(pt), and Malay(my).

### 4.2. In-House Ecommerce Benchmark Evaluation

We construct our in-house ecommerce benchmark based on real internal business scenarios. It includes 5 domains and 17 tasks, covering core ecommerce applications such as product guidance, after-sales issues, search intent understanding, product recommendation, and product title optimization. Table 2 summarizes performance over five domains (Ecom QA, Shopping Concept, User Understanding, Shopping Reasoning, Ecom Generation). **CompassMax-V3-Thinking delivers the top macro average** and is exceptionally strong on operational tasks that matter in production: *Product Recommendation* peaks at **94.58**, while *After-Sales* remains consistently high, evidencing robust decision quality under noisy, real-world inputs. These gains indicate that the reasoning improvements translate directly into higher-quality e-commerce actions, not just higher offline scores.

### 4.3. In-house Multilingual Benchmark Evaluation

We sample a set of representative and discriminative questions from our internally constructed general and ecommerce evaluation datasets. Local operations teams from various countries translate the

Table 2 | In-House Ecom Dataset Evaluation

| domain | Task | CompassMax-V3-Thinking | CompassMax-V3 | GPT-4o | GPT5-Thinking (medium) | GLM4.5 | Qwen3-235B-A22B | gemini-2.5-pro | Deepseek R1 |
|---|---|---|---|---|---|---|---|---|---|
| Ecom QA | Shopping Guide | 96.14 | 95.24 | 87.26 | 96.82 | 84.62 | 87.69 | 87.90 | 91.01 |
| | Brand Knowledge | 80.67 | 81.67 | 67.00 | 69.00 | 71.33 | 67.00 | 77.33 | 71.38 |
| | After Sales issue | 98.48 | 98.72 | 90.65 | 91.37 | 94.48 | 87.86 | 97.36 | 98.8 |
| Shopping Concept | Product classification | 82.33 | 83.00 | 75.00 | 70.33 | 73.33 | 69.67 | 72.67 | 72.00 |
| | Product information validation | 90.33 | 89.67 | 76.33 | 74.00 | 80.00 | 82.67 | 77.00 | 78.33 |
| | E-commerce attribute extraction | 88.00 | 86.33 | 81.33 | 81.33 | 78.67 | 75.67 | 82.67 | 80.78 |
| User Understanding | Query relevance judgment | 89.00 | 90.33 | 81.00 | 89.67 | 86.67 | 88.67 | 89.33 | 84.33 |
| | Seller violation detection | 82.33 | 81.67 | 82.00 | 78.67 | 80.00 | 79.00 | 82.00 | 82.00 |
| | Search intent understanding | 80.19 | 79.51 | 75.51 | 77.89 | 71.77 | 74.49 | 73.81 | 79.25 |
| | Search intent classification | 76.71 | 78.5 | 67.11 | 70.25 | 65.55 | 66.67 | 68.9 | 68.83 |
| Shopping Reasoning | Product information reasoning | 82.45 | 83.12 | 64.31 | 73.47 | 74.07 | 76.43 | 81.82 | 72.00 |
| | Tag relationship inference | 74.33 | 67.00 | 62.67 | 58.33 | 58.00 | 61.00 | 61.67 | 60.00 |
| | Product recommendation | 94.58 | 93.46 | 70.24 | 84.27 | 84.52 | 90.65 | 89.46 | 90.72 |
| | Title violation detection | 89.00 | 88.67 | 76.67 | 85.67 | 83.00 | 85.67 | 81.00 | 82.67 |
| Ecom Generation | Product ad rewriting | 90.67 | 90.00 | 83.67 | 79.33 | 80.67 | 84.67 | 90.33 | 81.94 |
| | E-commerce title optimization | 71.50 | 71.50 | 66.67 | 69.67 | 65.50 | 68.17 | 68.17 | 71.05 |
| | Product name rewriting | 91.67 | 89.00 | 61.67 | 46.8 | 69.00 | 84.67 | 93.67 | 79.55 |
| **Average** | Average | 85.79 | 85.14 | 74.65 | 76.29 | 76.54 | 78.27 | 80.89 | 79.10 |

English content into multiple languages, building our multilingual benchmark.

Table 3 | In-House Multilingual Datasets Evaluation

| Language | CompassMax-V3-Thinking | CompassMax-V3 | GPT-4o | GPT-5-Thinking(medium) | GLM4.5 | Qwen3-235B-A22B | gemini-2.5-pro | Deepseek R1 |
|---|---|---|---|---|---|---|---|---|
| en | 88.07 | 87.96 | 87.31 | 90.15 | 88.51 | 88.73 | 89.39 | 85.96 |
| tw | 85.64 | 83.08 | 88.32 | 89.58 | 88.11 | 88.44 | 88.84 | 86.38 |
| id | 87.89 | 88.71 | 80.02 | 88.08 | 85.31 | 86.76 | 86.62 | 85.08 |
| my | 86.56 | 84.63 | 86.17 | 87.51 | 86.61 | 87.31 | 88.28 | 83.39 |
| pt | 86.84 | 84.54 | 85.84 | 87.67 | 87.82 | 87.46 | 89.13 | 84.02 |
| th | 86.04 | 85.66 | 85.07 | 85.96 | 82.12 | 84.83 | 86.55 | 83.73 |
| vi | 84.75 | 84.38 | 80.99 | 83.97 | 84.09 | 83.26 | 85.76 | 84.34 |
| **SEA Average** | 86.41 | 85.58 | 83.62 | 86.64 | 85.19 | 85.92 | 87.27 | 84.11 |

Table 3 shows performance over seven Southeast Asian languages. CompassMax-V3-Thinking achieves a **SEA Avg 86.41** with tight, balanced language-wise performance (e.g., EN 88.07; ID 87.89), matching the stability profile required for cross-market rollouts and narrowing gaps with larger frontier models without sacrificing consistency in low-resource languages.

### 4.4. In-house General Benchmark Evaluation

Table 4 | General Ability Evaluation Based on In-House Benchmark

| Domain | CompassMax-V3-Thinking | CompassMax-V3 | Qwen3-235B-A22B(think) | gemini-2.5-pro(think) | GPT-5-Thinking(medium) | DeepSeek-R1 |
|---|---|---|---|---|---|---|
| Knowledge & Comprehension | 62.07 | 56.48 | 61.39 | 65.87 | 68.82 | 62.70 |
| Creative Generation | 77.75 | 62.34 | 73.51 | 74.63 | 75.95 | 74.55 |
| Reasoning | 74.54 | 62.73 | 74.54 | 78.31 | 80.94 | 69.91 |
| Code | 76.83 | 67.94 | 79.87 | 82.06 | 85.08 | 80.16 |
| Safety | 88.86 | 72.95 | 80.19 | 86.25 | 86.10 | 88.57 |
| **Average** | 76.01 | 64.49 | 73.90 | 77.42 | 79.39 | 75.19 |

Table 4 shows the results over five domains. CompassMax-V3-Thinking raises the overall average to **76.01** (vs. 64.49 for CompassMax-V3), a double-digit lift driven by sizeable gains in *Creative Generation*, *Safety*, and *Code*. The profile indicates stronger structured thinking and reliability, not only better factual recall—key for production-facing assistants.

### 4.5. Open-source ARC Benchmark Evaluation

Table 5 shows performance over ten popular open-source benchmarks. CompassMax-V3-Thinking shows strong reasoning+format discipline (*IFeval* **85.40**), competitive high-level knowledge (*GPQA-Diamond* **68.69**), and substantial math improvements (*AIME24/25* **83.30/80.00**; *HMMT* **46.70**). Coding quality is production-ready (*HumanEval* **98.17**; *MBPP* **73.54**). On agentic tasks, it performs robustly across *BFCL* variants (notably improving the multi-turn setting to **19.50**), evidencing better planning and recovery in longer tool chains.

Table 5 | ARC Ability Evaluation Based on Open-Source Benchmark

| Domain | Benchmark | **CompassMax-V3-Thinking** | CompassMax-V3 | DeepSeek-R1 |
| --- | --- | --- | --- | --- |
| *Alignment* | IFeval$_{\text{(prompt strict)}}$ | 85.40 | 83.55 | 84.84 |
| *General QA* | MMLU-Redux$_{\text{(EM)}}$ | 87.23 | 42.87 | 88.17 |
| | GPQA Diamond$_{\text{(EM)}}$ | 68.69 | 44.44 | 69.19 |
| *Agent* | BFCL_AST_NON_LIVE | 83.73 | 86.46 | 87.36 |
| | BFCL_AST_LIVE | 79.57 | 73.20 | 80.61 |
| | BFCL_MULTI_TURN_LIVE | 19.50 | 12.00 | 12.38 |
| *Reasoning* | AIME24$_{\text{(Pass@1)}}$ | 83.30 | 26.67 | 79.80 |
| | AIME25$_{\text{(Pass@1)}}$ | 80.00 | 13.33 | 70.00 |
| | HMMT$_{\text{(Pass@1)}}$ | 46.70 | 11.52 | 41.67 |
| | Zebralogic$_{\text{(Acc)}}$ | 77.80 | 18.50 | 78.70 |
| *Coding* | HumanEval$_{\text{(Pass@1)}}$ | 98.17 | 84.76 | 96.95 |
| | MBPP$_{\text{(Pass@1)}}$ | 73.54 | 67.70 | 77.43 |

## 5. Conclusion

We presented a unified RL framework for LongCoT models that combines multi-stage zero-variance elimination, entropy-adaptive optimization, stable reward modeling, and high-throughput rollout engineering. These techniques jointly produce a more efficient and reliable training pipeline for large reasoning models.

Empirically, the resulting CompassMax-V3-Thinking model shows that the algorithmic gains translate directly into practical capability improvements. Internal e-commerce evaluations report clear jumps in decision quality—for example, recommendation tasks reach above 94, and after-sales reasoning remains consistently high under noisy inputs—indicating that enhanced reasoning yields more reliable downstream actions. Multilingual testing shows balanced cross-market behavior with averages in the mid-80s, meeting deployment-level stability requirements. General-domain evaluations also exhibit substantial improvements, raising the overall average from the mid-60s to over 75, driven by gains in creative reasoning, safety, and code performance. Public benchmarks further validate these trends, with strong reasoning-format compliance, sizeable math improvements, and near-perfect coding accuracy on HumanEval.

Taken together, these results show that our RL pipeline fosters a model that is not only more capable in long-context reasoning but also more stable and production-ready across domains. The framework provides a scalable foundation for training next-generation agentic and domain-adaptive reasoning systems.

## Contributors

*Authors are listed alphabetically by the first name.*
Anxiang Zeng
Haibo Zhang
Hailing Zhang
Kaixiang Mo

Liang Yao

Ling Hu

Long Zhang

Shuman Liu

Shuyi Xie

Yanshi Li

Yizhang Chen

Yuepeng Sheng

Yuwei Huang

Zhaochen Xu

Zhiqiang Zhou

Ziqin Liew